# FishRecGAN: An End to End GAN Based Network for Fisheye Rectification and Calibration

**Xin Shen**    xinshen@andrew.cmu.edu
*Carnegie Mellon University*
*Pittsburgh, PA 15213, USA*

**Kyungdon Joo**    kjoo@andrew.cmu.edu
*Carnegie Mellon University*
*Pittsburgh, PA 15213, USA*

**Jean Oh**    jeanoh@andrew.cmu.edu
*Carnegie Mellon University*
*Pittsburgh, PA 15213, USA*

**Corresponding Author:** Xin Shen

## Abstract

We propose an end-to-end deep learning approach to rectify fisheye images and simultaneously calibrate camera intrinsic and distortion parameters. Our method consists of two parts: a Quick Image Rectification Module developed with a Pix2Pix GAN and Wasserstein GAN (W-Pix2PixGAN), and a Calibration Module with a CNN architecture. Our Quick Rectification Network performs robust rectification with good resolution, making it suitable for constant calibration in camera-based surveillance equipment. To achieve high-quality calibration, we use the straightened output from the Quick Rectification Module as a guidance-like semantic feature map for the Calibration Module to learn the geometric relationship between the straightened feature and the distorted feature. We train and validate our method with a large synthesized dataset labeled with well-simulated parameters applied to a perspective image dataset. Our solution has achieved robust performance in high-resolution with a significant PSNR value of 22.343. [1]

## 1. INTRODUCTION

Fisheye cameras have become popular in robotics-related industries due to their large field of view, but they introduce severe distortion and nonlinearity among pixels. To address this issue, the industry relies on traditional fisheye camera calibration [1, 2], which requires an individual to hold a checkerboard in front of the camera and take pictures with different poses. These pictures are then fed into a geometric algorithm to calibrate the camera intrinsic and distortion parameters. Many existed works can be operated through the OpenCV library [3] following the traditional method [4]. However, this process requires a significant amount of human labor.

---

[1] This work was done while authors attended Carnegie Mellon University at 2020. To communicate, please contact author Xin Shen via xinshen@alumni.cmu.edu

## 1.1 Motivation

The conventional calibration method exhibits inconsistencies and necessitates significant human involvement, thereby introducing the potential for human errors. Furthermore, it lacks the capability to promptly rectify camera settings in real-time without pre-calibration. Additionally, it relies on specific equipment and mandates the utilization of a compatible camera to capture images with a checkerboard for optimization purposes. Consequently, our objective is to present an algorithmic solution that is independent of human intervention, ensuring consistency and efficiency. Our proposed algorithm aims to enable real-time rectification for camera surveillance operations, which necessitate continuous adjustments during their execution.

## 1.2 Our Contribution

In this paper, we proposed an enhanced approach to construct an end- to- end multi-contextual network architecture consisted of GANs and CNNs. The architecture can be found in the FIGURE 1. Specifically, we make the following major contributions:

- We proposed an end- to- end GAN based multi- contextual network to better learn the geometric mapping relationship between the distorted nonlinear pixels (fisheye images) and the rectified linear pixels (fisheye- rectification) as an enhancement for one of the previous state- of- art work [5]. We developed a novel training algorithm for the Pix2Pix GAN model [6] by integrating Wasserstein GAN's (WGAN) [7] approach. This allowed the model to rectify a fisheye image to its corresponding straightened image pair with high resolution (avg. PSNR 22.343 4.2) without the need for traditional geometric algorithm via the calibrated parameters, which is computationally expensive. The single GAN model provided the ability to rectify fisheye images for surveillance equipment that requires constant calibration.

- We synthesized a large- scale dataset consist of fisheye- image- straightened- image pairs with the corresponding parameters. This dataset contains both clear structural framework and weak structural image pairs with a well- simulated distortion parameters which provides consistency for a deep neural network to learn.

## 1.3 Overview of Our Approach

We use Generative Adversarial Networks (GAN) [8] to solve the problem of fisheye image rectification, which involves finding the mapping function between nonlinearity and linearity among pixels. Compared to many traditional computer vision based algorithms to infer an object's geometric conditions [1, 2, 9, 10], GAN is advantageous because it can achieve real-time performance during inference with its lightweight architecture. Specifically, we use the Pix2Pix GAN to solve the direct mapping problem from fisheye image to perspective image. However, Pix2Pix GAN struggles with high differences between distributions, such as those found in fisheye images, making it difficult to learn in a high-level manifold. To address this, we incorporate the Wasserstein GAN (WGAN) learning algorithm, using the Earth Mover (EM) distance to provide a continuous learning curve. Our proposed W-Pix2PixGAN model achieves a high-resolution direct rectification from fisheye image to its corresponding perspective image FIGURE 2, with an average PSNR score of 22.343.

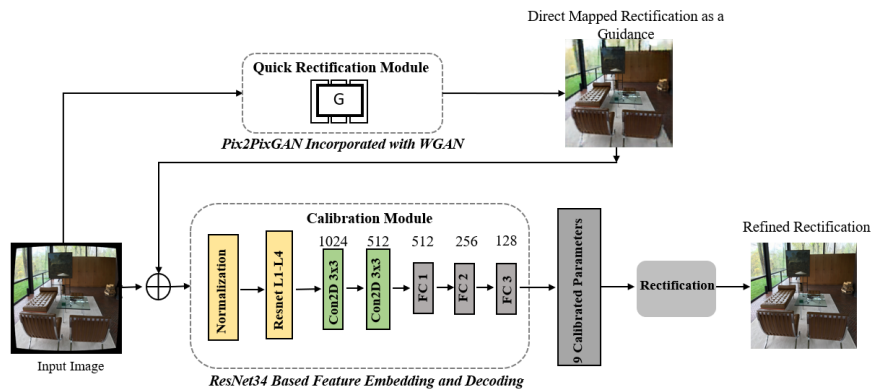

Figure 1: The model comprises three key components: the Quick Rectification Module, the Calibration Module, and the Rectification Layer. The Quick Rectification Module, based on an enhanced Wasserstein GAN and Pix2Pix GAN, generates a ground-truth-like semantic guidance and performs real-time preliminary rectification. The Calibration Module employs a ResNet-based CNN architecture, utilizing the concatenated feature to extract pixel relationships and calibrate parameters for curved-to-straightened pixel mapping. The rectification layer utilizes the obtained distortion parameters to perform image rectification.

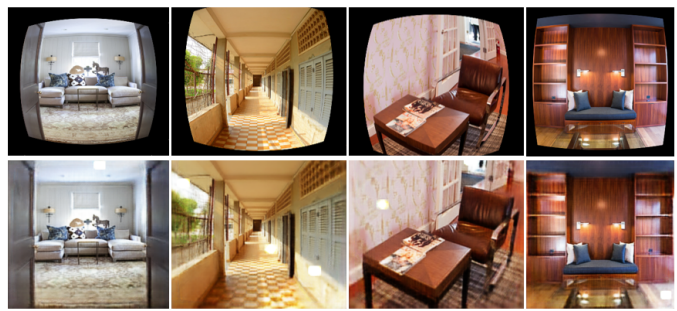

Figure 2: The quick rectification performed by our WGAN enhanced Pix2Pix GAN model. **1st row**: original synthesized fisheye images; **2nd Row**: rectified images

In many industrial applications, calibrated camera parameters are required for further use. While convolutional neural networks (CNNs) have been explored for predicting visual-based camera parameters [11], a simple feed-forward CNN architecture is often more suitable for subtle visual classification and object detection [12, 13]. However, for regression tasks such as predicting camera parameters, a deep learning model requires more geometric information constraints than just a single raw fisheye image as the sole input feature. To address this issue, we focused on creating a strong inter-pixel relationship feature map for a convolution network to learn the mapping function between given features and the parameters for regression, following a similar idea presented in [14]. In line with Xue, Zhucun and colleagues, we used the lines detected on the raw fisheye image from the line detection network as a "semantic guidance" concatenated with the raw fisheye image to create a new feature map for a ResNet based model to learn [5, 15]. We proposed an assumption that this would help to enhance the performance more than the previous work. Thus, unlike this previous work, we concatenate the output from the W-Pix2PixGAN, which is already a ground- truth- alike feature, to the raw fisheye image. By doing so, we are able to create an inter- relationship between any curved structures in the fisheye image and the corresponding straightened ones which are how the curved supposed to be rectified. Then this new feature map is fed into the an similar Parameter-Calibration Module architecture as the previous work to perform the regression/ calibration, which is to predict 9 distortion parameters.

## 2. RELATED WORK

In 2019, Wuhan University proposed the "Multi-contextual Network" approach, which introduces a "Guidance-alike" semantic feature map generated from a CNN and concatenated with the original fisheye image for enhanced learning [16]. Previous work in 2019 by Xue et al. [5] used a line detection network to highlight distorted lines in fisheye images and concatenated them with the original image to create a feature map containing more geometric information for a ResNet-based regression network to learn. The architecture consists of a line detection network, a calibration module with ResNet, and a traditional geometric rectification method to take in the calibrated parameters for rectification.

While the use of distorted lines as guidance for introducing more geometric information on the pixel level is an innovative idea, it falls short in providing a real-time solution for fisheye image rectification. The pipeline still requires simulating the calibration process to obtain distortion parameters, and during inference, the multi-contextual network requires too much computation to run in real-time. Additionally, using distorted lines as guidance may have limitations in detecting well-structured lines in non-line sensitive input images, which can reduce the quality of the semantic feature.

Our objective in this research study was to enhance the existing work by focusing specifically on the semantic generation aspect within the pipeline. We operated under the assumption that improving the quality of semantic features would yield superior calibration outcomes. Notably, recent studies [17–19] have demonstrated that employing generic semantic guidance leads to significant improvements in model performance for regression and classification tasks. In order to further enrich the information provided to the network, we proposed the incorporation of a ground-truth-like feature, referred to as the corresponding perspective image. This feature not only rectifies all lines within the image but also exhibits linear pixel patterns, thereby enhancing the effectiveness of guidance in the calibration process. To accomplish this, we leveraged generative adversarial networks (GANs) and trained them using a pair of distorted and perspective images. The GAN model facilitated the generation of a ground-truth-like feature, enabling rapid rectification through the use of the generator alone. We integrated the training algorithm of Wasserstein GAN (WGAN) and introduced modifications to the original loss function of Pix2Pix GAN, incorporating the Earth Mover's (EM) distance, thereby enhancing the GAN model's performance and yielding high-resolution outputs.

## 3. TECHNICAL APPROACH

To validate our approach and assumption, we aimed to replicate the previous work as closely as possible, with the exception of replacing the original line detection network with the W-Pix2PixGAN for semantic generation. However, the previous work's authors did not publicly share their implementation, so we developed a similar dataset by implementing a fisheye-image-synthesis algorithm with the same camera model mentioned in the paper. Through simulating fisheye-effect-synthesis, we identified the parameters needed to generate a fisheye image similar to the previous work. During training and inference, we followed the same pipeline as the previous work, randomly selecting four distortion parameter sets out of the total twelve to synthesize the fisheye image. We replicated a similar Calibration Module architecture using ResNet34 as the backbone. By changing only the first part of the network, we conducted a fair comparison to determine which semantic generation model provided better guidance semantics.

### 3.1 Data Synthesis and General Fisheye Camera Model

To train the model, we synthesized our datasets by distorting a perspective image using a general polynomial projection model[20]. With a given normal perspective pinhole camera, a point $\mathbb{P} := \{X, Y, Z\} \in \mathbb{R}^3$ in the world frame can be projected onto the image frame $\mathbb{P}_i := \{u, v\} \in \mathbb{R}^2$ in the following transformation using the camera intrinsic matrix. See Appendix A for detailed mathematical models and derivations.

### 3.2 Parameter Selection and Simulation

Since the authors of the previous work, Zhucun Xue and et al [5], did not provide the distortion parameters they used, we picked up our own parameters which yield the similar fisheye distortion effect shown in their work.

To be consistent with the previous work, we generated our synthesized dataset by artificially adding distortion upon the WireFrame Dataset [21] with randomly selecting 4 distortion parameters out of 12. In FIGURE 4, we list several samples of the distortion effect, such as full- frame fisheye image, minor- distortion image, drum- fisheye image, and full- circle image.

Machine learning algorithms often face the difficulty of learning a non-deterministic and inconsistent mapping function. The problem of generating a fisheye effect from a given perspective image is particularly challenging, given the many random combinations of the nine distortion parameters involved. Blindly and randomly selecting parameter combinations can make it difficult for the network to learn the transformation pattern. To overcome this issue, we conducted a simulation process that changed one parameter at a time while ruling out the others and observed the physical effect of each parameter. We varied each parameter from -0.9 to 1 and visualized the effects of changing each $k_i$ at the same level in FIGURE 3. Upon observation, we discovered that:

- $k_1$ is doing the major contribution which has a sensitive and significant effect on both the center and the edge of a given perspective image.

- $k_2$ and $k_3$ have a less sensitive effect on distorting an image and both have a slight impact on the center of the image.

- $k_4$ and $k_5$ almost have no effect in distorting the center pixels while both have a slight and non- sensitive effect on the edge.

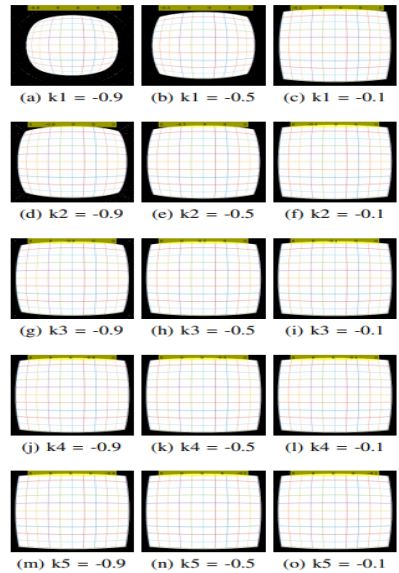

Figure 3: Shows the distortion parameters' simulation to figure out the proper ones for fisheye image generation

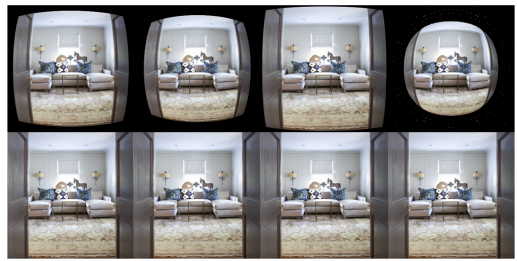

Figure 4: A sample showing the synthesized fisheye- perspective image pair

From the $1^{st}$ row of FIGURE 3, we have found that by only changing $k_1$ could we obtain a similar visual distortion effect as the previous works. However, in order to increase the model's generaliza-tion ability and meanwhile to keep the parameters consistency, we chose $k_2, k_3, k_3, k4, k_5$ to be as simple as possible but not to be 0. In FIGURE 4 shows a set of samples of our synthesized dataset with 12 different fisheye distortion effect.

### 3.3 Deep Rectification and Calibration Network

In this section, we mainly exploit the details of the two major modules of our model, namely the W-Pix2PixGAN model and the Calibration Module with Resnet34 as the backbone. Meanwhile, we will introduce the training scheme and the loss function designed.

As shown in FIGURE 1, our full model is mainly consisted of two major deep neural networks. The first one is the Rectification Module consisted of W-Pix2PixGAN model to perform a preliminary and quick rectification with a given fisheye image. The second is the Calibration Module with ResNet34 as the backbone; we built this module in the similar architecture, such as filter sizes and convolutional layers designs, as much similar as possible to the previous work [5]. This module is designed to perform the estimation of the 9 important parameters including the distortion parameters $K_d$ as universal regressor.

Given a RGB fisheye image $I$ with size of $H \times W$, a rectified semantic map $\mathbb{H} \in \mathbb{R}^{\mathbb{H} \times \mathbb{W}}$ is generated from the Rectification Module, and then this semantic feature is used as a guidance to be concatenated with the original fisheye image to create a new feature $F$. This new feature is then fed into the Calibration Module to learn the inner-pixel relationship between the curved lines and the corresponding rectified line in a high manifold, and finally to learn the 9 parameters through a multi-layer perception network. Thus, for our model, every training data sample contians: (1) a fisheye image $I$, (2) the ground truth of the corresponding rectified image map $H$, (3) the ground truth of the distortion parameters $K_d$.

We used the architecture of the Pix2PixGAN for the W-Pix2PixGAN model, which includes the U-Net structure for the generator and the patch-discriminator structure. However, since we needed the model to learn a mapping between two distributions, the fisheye distribution and the rectified distribution, we used Instance-Normalization instead of Batch-Normalization. Initially, training the Pix2PixGAN model was challenging due to the inherent limitation of the GAN's objective function, which minimized the difference between the Jensen–Shannon divergence between the real and fake distributions and a constant value. However, this led to zero-overlapping in the high manifold and a lack of learning. To resolve this, we used a $16 \times 16$ patch design and modified the discriminator's architecture to use a linear layer instead of a sigmoid layer, so that the output is a regression used as the GAN loss. Overall, we used the original structural design of the generator and the $16 \times 16$ patch-discriminator design with four convolutional layers.

**Calibration Module.** In order to validate our assumption that by using GAN as the semantic generation part, we needed to control variables. Thus, we tried to follow the original architecture of the previous work's design as much as possible; however, most architecture details were not clearly indicated.This module is trimmed to estimate the distortion parameters from the concatenated features. As mentioned above, the input feature for this module is the concatenation of the rectification map $H$ and the raw fisheye image $I$ with the size of $H \times W \times 6$. As shown in the FIGURE 1, we applied a 4-level ResNet-34 [15] as the backbone for this module. A high-level dense feature map out from the L1- L4 ResNet is then fed to 2 other convolutional layers to be introduced with more nonlinearity with LeakyReLU activation. A 3- layer fully connected (FC) layers are then connected after the last convolutional layer. In order to restrict the model's learning behavior, we did not introduce any dropouts within the FC layers, and as this is a regression problem, we used all linear activation within the FC layers. The last FC layer is used to predict a 9- D vector representing the distortion parameters denoted by $K_d$.

Table 1: The summary of the receptive field in each convolutional layer. As shown in the table, with this design we could achive a high receptive field up to 70 at the last convolutional layer

| Layer Number | Kernel Information | Receptive Field |
|---|---|---|
| conv_layer1 | [4 x 4, 64], s = 2, p = 1 | 4 |
| conv_layer2 | [4 x 4, 128], s = 2, p = 1 | 10 |
| conv_layer3 | [4 x 4, 256], s = 2, p = 1 | 22 |
| conv_layer4 | [4 x 4, 512], s = 2, p = 1 | 46 |
| conv_layer5 | [4 x 4, 1], s = 2, p =1 | 70 |

**Rectification Layer.** In this module, we followed geometric model in Eq. (10) to iteratively remove the distortion parameters predicted using bi-linear interpolation.

$$P_d = \tau(p, K_d) = \begin{bmatrix} u_0 \\ v_0 \end{bmatrix} + \frac{r(\theta)p}{\|p\|_2} \qquad (1)$$

, where the pixel coordinate in the rectified image is $\mathcal{P} = (x, y)$, and the pixel coordinate in the fisheye image is $\mathcal{P}_\mathcal{D} = (x_d, y_d)$.

### 3.4 Loss Function and Training Scheme

In our network, which performs both quick rectification by W-Pix2PixGAN and distortion parameter calibration by a ResNet-based CNN, we performed supervised training for both modules. To pre-train the GAN, we provided a pair of images: a fisheye image denoted as $Real_A$ and a ground truth picture of the rectified perspective image denoted as $Real_B$. The learning objective of the GAN was to learn a direct mapping between the fisheye image and the generated rectified image, denoted as $Fake_B$. Following the scheme in FIGURE 1, we next used the generated rectified image as guidance and concatenated it with the raw fisheye image to create a new feature map, which was then fed into the Calibration Module. This learning was supervised by the ground truth of the 9 distortion parameters $K_d$, and in turn, this network was trained to perform a universal regression to predict the corresponding parameters using the concatenated feature map.

We integrated the original Pix2PixGAN model [6] with the Wasserstein GAN's idea of EM distance [7] to achieve a continuous GAN loss for training. This enabled our model to successfully learn the mapping function between two significantly different pixel distributions. We modified the original MSE loss between the probability output distribution from the discriminator and the truth distribution (either all ones or all zeros, representing being real and being fake, respectively) to the EM distance by removing the last sigmoid layer of the discriminator, $f_w$. The input fed into the discriminator was identical to the original Pix2PixGAN's design, where we concatenated $Real_A$ to $Real_B$ as a new distribution, $P_r$, to train the discriminator to recognize the real distribution.

Similarly, we concatenated $Real_A$ to $Fake_B$ as a new distribution in $P_g$ to train the discriminator to recognize the fake distribution. The expectation of the output distribution from the discriminator was directly treated as the GAN loss. The discriminator loss and the generator loss are shown in Eq. 2 and Eq. 3, respectively.

$$\mathcal{L}_D = \mathbb{E}_{x \in P_r}[f_w(x)] - \mathbb{E}_{x \in P_g}[f_w(x)] \tag{2}$$

$$\mathcal{L}_G = -\mathbb{E}_{x \in P_g}[f_w(x)] \tag{3}$$

The previous loss function designed enlightened by Wasserstain GAN helps a continuous learning curve when the Pix2PixGAN is faced with 2 significantly different distribution; however, meanwhile, the generator's role is not only to fool the discriminator but also to generate an output as closer to the ground truth as possible. Thus, we also utilized the orginal pixel loss by using the $L_1$ loss between the generator's output and the ground truth, shown in Eq. 4.

$$\mathcal{L}_{L1(G)} = \mathbb{E}\|y - G(x)\|_1 \tag{4}$$

Overall, our final objective is shown in Eq.5.

$$G^\star = \underset{D}{argmax}\mathcal{L}_D + \underset{G}{argmin}\mathcal{L}_G + \lambda\mathcal{L}_{L1(G)} \tag{5}$$

Lastly, the pseudo code of our training algorithm can be found below:

---

**Algorithm 1:** Training Algorithm for W-Pix2PixGAN

---

**Require:** $\alpha$, the learning rate. c, the weight clipping parameter. m, the batch size. n, how many more iterations to train discriminator more

**while** *Within training epochs* **do**
    **for** *t = 0, ..., n* **do**
        Sample $\{x_A^{(i)}\}_{i=1}^m$ from fisheye data;
        Sample $\{x_B^{(i)}\}_{i=1}^m$ from perspective data;
        $g_B = f_\theta(x_A)$;
        $d_A^{(i)} = cat(x_A^{(i)}, x_B^{(i)})$;
        $d_B^{(i)} = cat(x_A^{(i)}, g_B^{(i)})$;
        $G_w \leftarrow \nabla_w[\frac{1}{m}\sum_1^m f_w(d_A^{(i)}) - \frac{1}{m}\sum_1^m f_w(d_B^{(i)})]$;
        $w \leftarrow w + \alpha \cdot RMSProp(w, G_d)$;
        $w \leftarrow clip(w, -c, c)$;
    **end**
    Sample $\{z_A^{(i)}\}_{i=1}^m$ from fisheye data;
    $g_B^{(i)} = G_\theta(z_A^{(i)})$;
    $d_B^{(i)} = cat(z_A^{(i)}, g_B^{(i)})$;
    $G_\theta \leftarrow \nabla_\theta[-\frac{1}{m}\sum_1^m f_w(d_B^{(i)}) + \lambda \cdot \mathcal{L}_{L1(G)}]$;
    $\theta \leftarrow \theta - \alpha \cdot RMSProp(\theta, G_\theta)$;
**end**

---

The training was done on Nivida 1080Ti GPU device with 500 epochs and the learning rate was set to decay dynamically with respect to the validation performance using PyTorch.

**The Training of Calibration Module.** In this module, the learning goal is to build a universal regressor to predict the 9 distortion parameters $K_d$. Thus, ideally, we perform a L2 loss upon the prediction against the ground truth $K_{gt}$. However, as shown in FIGURE 3, we have found out that among all 9 parameters, $K_1$ is making the significant impact on the distortion effect both on the center and on the edge of the image. Thus, we performed a weighted L2 loss which emphasizes on $K_1$ more with a parameter $\beta$.

$$\mathcal{L}_{L2} = \frac{1}{9}[\beta \cdot (K_g(1) - K_{gt}(1))^2 + \sum_{i=2}^{9}(K_g(i) - K_{gt}(i))^2] \tag{6}$$

Similarly, this training was implemented with PyTorch using Nvidia TITAN GPU device for 500 epochs, and the learning rate was set to decay dynamically with respect to the validation performance.

## 4. EXPERIMENT AND EVALUATION

### 4.1 Implementation Details

We randomly selected 4 out of the total 12 distortion parameters and applied them to the WireFrame dataset, creating 20,000 training samples and 1,848 test samples. We trained the Rectification Module (W-Pix2PixGAN) for 500 epochs using the training scheme outlined in Section 3.4. We used an initial discriminator learning rate of $Lr_D := 0.0009$, an initial generator learning rate of $Lr_G := 0.0001$, and a batch size of 32. We also allowed for dynamic learning rate decay with respect to the validation performance to refine the GAN model output resolution during training.

We concatenated the output from the W-Pix2PixGAN with the raw fisheye image to create a new feature map to train the calibration network for 500 epochs with an initial learning rate of 0.001 and a batch size of 16. We also allowed for dynamic learning rate decay with respect to the validation loss. We used $\beta$, the weighted penalty upon distortion $K_1$, as 32. Finally, during inference, we loaded the best performing weights for both models and performed quick rectification, followed by concatenation and calibration, and then calibration and fine rectification sequentially.

### 4.2 Evaluation Details

As the authors of the previous model have not yet published their code, we were unable to access their line detection module. To assess the impact of our approach, which replaces the line detection module with W-Pix2PixGAN, we assumed that our Calibration Module operates similarly to that of the previous work. As a measure of the quality of the guidance feature map, we concatenated the ground truth of distorted fisheye lines used in the previous work to the raw fisheye image. We then compared the rectified fisheye image using our approach to that of the previous work, using the predicted distortion parameters $K_{Dpred}$. To evaluate the quality of the rectified image, we used the peak signal to noise ratio (PSNR) and the structure similarity index (SSIM) [22], following the evaluation metrics used in the previous work [5]. To assess the fairness of this comparison, we also compared the PSNR and SSIM scores of the baseline output to the ground truth of the perspective image. We then used these metrics to evaluate the performance of our W-Pix2PixGAN model for quick rectification. Finally, we used the distributions of the differences in PSNR and SSIM scores between the baseline output and our model's output to construct 95% confidence intervals, and checked whether 0 was within each interval.

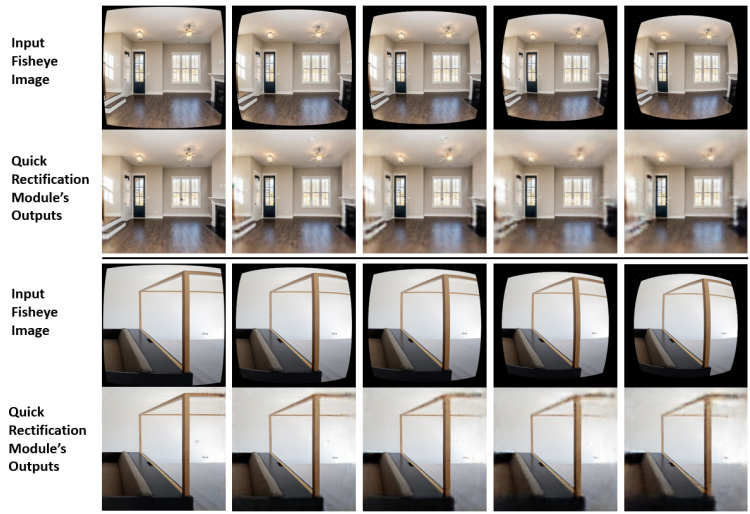

Figure 5: Our Quick Rectification Module demonstrates effective performance, rectifying curved structures back to straightened forms with high resolution.

## 5. EXPERIMENT RESULTS

### 5.1 The Performance by Quick Rectification Module

As one of our objectives is to provide a direct and quick rectification given any fisheye images without going through a calibration work either by human labor or through a computationally heavy calibration network with ResNet as the backbone, we put a lot of attention on refining our W-Pix2PixGAN model, and in turn we have provided the PSNR and SSIM calculation on the 1,848 test samples with 4 randomly selected distortion parameters applied. We then separately sampled out the quick rectification performed by GAN for distortions, such as, *minor distortion*, *drum- fisheye image*, *severe- drum- fisheye image*, *full- frame fisheye image*, and *severe-full- frame-fisheye iamge*. From the range of minor distortion to severe full- frame fisheye distortion, as shown in the FIGURE 5, our W-Pix2PixGAN model could perform a quick and high- resolution rectification work directly from a given fisheye image by learning a universal pixel- to- pixel mapping relationship.

In the TABLE 2, we show the summaries of the quality of PSNR and SSIM. Compared to the previous work's overall average PSNR 27.61 and SSIM 0.8746 via the full- pipeline rectification using the calibrated parameters, we can see that given that it is solely a quick rectification by image transferring performed by W-Pix2PixGAN, the quantitative performance by our GAN model solely is acceptable.

Table 2: The summary of the W-Pix2PixGAN's performance on each separated distortion set and on the full- dataset with randomly selected 4 distortions

|  | AVG. PSNR | AVG. SSIM |
|---|---|---|
| Minor Distortion | 27.7673 | 0.8733 |
| Full-frame Fisheye Distortion | 23.4372 | 0.7431 |
| Drum-fisheye Distortion | 24.357 | 0.7823 |
| Full Dataset with 4 Random Distortions | 22.343 | 0.7185 |

## 5.2 Full- pipeline Comparison to Previous Work

Following the evaluation protocol outlined in Section 4.2, we conducted an end-to-end rectification process, beginning with the input of raw fisheye images and proceeding to rectify them using predicted distortion parameters. The calibration network remained fixed throughout, with the only variation being the guidance semantic concatenated with the fisheye image, using both ground truth fisheye lines and the output from our GAN. However, our attempts to replicate the quantitative results of the previous work in terms of PSNR and SSIM, as presented in TABLE 3, were unsuccessful. This may be attributed to the fact that duplicating a specific neural network necessitates a more comprehensive understanding of its details, despite our efforts to recreate the calibration network based on the information provided in the previous work's publication.

Table 3: The summarized results comparing the averaged PSNR and SSIM by both using the ground- truth based baseline and our approach

|  | Average PSNR | Average SSIM |
|---|---|---|
| *Our Approach* | 23.4717 | 0.7344 |
| Baseline via Ground Truth | 23.4263 | 0.7326 |

However, both yielded a enhanced performance compared to solely using the Quick Rectification Module with a pair of very closed averaged result. By following the evaluation pipeline in Section. 4.2, we constructed a 95% confidence interval on the distribution of difference between the baseline model and our model for both PSNR and SSIM respectively shown in FIGURE 7.

The results presented in TABLE 4 demonstrate that we have obtained narrow confidence intervals for both PSNR and SSIM, encompassing a range where zero difference is observed. This indicates that our approach, which replaces the line detection model with W-Pix2PixGAN while utilizing the ground truth for distorted fisheye lines as an upper bound, does not exhibit a significant disparity. Thus, we consider our approach to be comparable to the previous work. Additionally, we provide a compilation of rectification performance achieved by utilizing the predicted distortion parameters below.

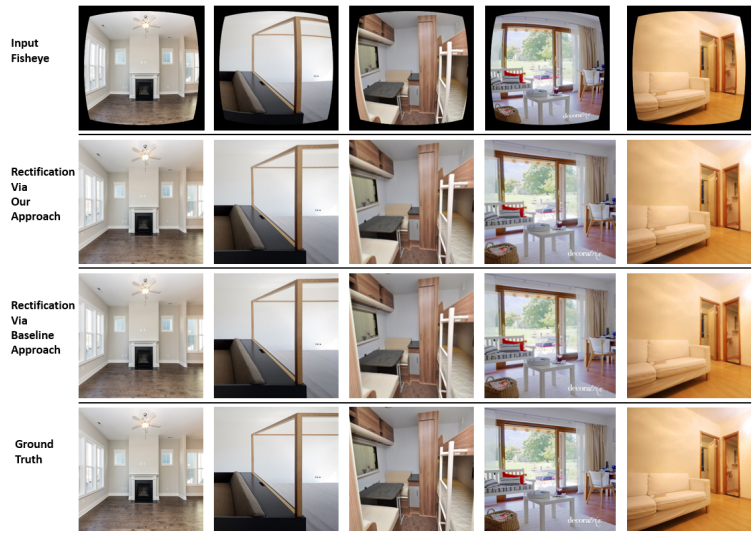

Figure 6: Experimental results demonstrate the rectification performance achieved through the predicted distortion parameters using both our proposed approach and the baseline method.

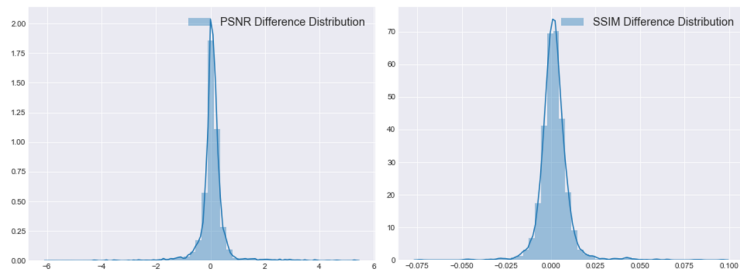

Figure 7: Shows the distribution of the SSIM difference between the baseline performance and our model's performance

Table 4: Confidence Interval at 95% significant level for PSNR and SSIM

|  | PSNR Difference | SSIM Difference |
|---|---|---|
| Confidence Interval at 95% | [-0.0369, 0.0538] | [-0.0016, 0.0018] |
| Marginal Error | ± 0.008479 | ± 0.00012 |

The results with structured dataset are promising. As mentioned in Section 4.2, this trial of experiment shows an unfairness for our approach because of the fact that we are comparing our method with the ground truth of the distorted lines but our approach still shows a statistically proved comparable result with a slight improvement based on TABLE 3. Meanwhile, the baseline approach via using line detection model shows a clear advantage of the obvious presence the lines of within distorted images and in turn is expected to perform well on structured dataset, such as the WireFrame dataset used in this experiment. However, the previous work's approach might show a limitation when faced with unstructured dataset where the edges of lines in one image is not clear to be detected, such as human faces. However, our GAN based approach is not limited by the nature of the dataset regardless of being structured or unstructured. Thus, this stimulates a further experiment on dataset such as CelebA [23] where the images might not include rich information in terms of lines for the baseline approach to exploit.

## 6. CONCLUSIONS

In this paper, we presented an enhanced approach for improving fisheye image calibration and rectification using a multi-contextual neural network. Our method incorporates a GAN-based semantic-guidance generator, which provides a ResNet-based calibration network with a ground-truth-like semantic feature, enabling end-to-end automatic fisheye image rectification from a single input image. Due to the unavailability of detailed implementation information regarding the secondary calibration network from the previous work, we were unable to replicate their exact experimental results. However, we anticipate performing a thorough evaluation once the authors of the previous work release their implementation. Statistically, we have demonstrated that our approach does not exhibit a significant difference compared to using the ground truth of distorted fisheye lines as an upper bound for the output of the previous work's line detection module. This validates our assumption that improved guidance leads to better calibration, as the direct utilization of a ground-truth-like feature proves advantageous over detecting distorted lines as the guidance. Consequently, refining the GAN model becomes crucial. Furthermore, the baseline approach may encounter challenges when confronted with unstructured images such as human faces. Hence, for future work, we plan to apply our approach's pipeline to unstructured datasets, such as Celeb-A [23]. Recent advancements in transformer-based structures have demonstrated significant improvements in visual reconstruction and regression tasks [24–26]. In our future research, we aim to explore the potential of utilizing these advanced transformer models for our Calibration Module, comparing their performance against our current *ResNest34* architecture. In summary, our future research will concentrate on two main objectives: (1) improving our model architecture through the incorporation of contemporary advancements in transformers, and (2) reassessing the complete pipeline when the authors of the previous work make their implementations publicly available.

## Appendix A. Data Synthesis Mathematical Modeling

We list the detailed mathematical modeling to generate synthetic data in this section.

Below shows a general camera model.

$$\begin{bmatrix} u \\ v \\ 1 \end{bmatrix} = \begin{bmatrix} f_x & 0 & c_x \\ 0 & f_y & c_y \\ 0 & 0 & 1 \end{bmatrix} \begin{bmatrix} X_c \\ Y_c \\ Z_c \end{bmatrix} \quad (7)$$

, where the $\mathbb{P}_c = \{X_c, Y_c, Z_c\} \in \mathbb{R}^3$ is the coordinates transformed from the world coordinate onto the camera frame using the following transformation.

$$\begin{bmatrix} X_c \\ Y_c \\ Z_c \end{bmatrix} = \begin{bmatrix} r_{11} & r_{12} & r_{13} \\ r_{21} & f_{22} & r_{23} \\ r_{31} & r_{32} & r_{33} \end{bmatrix} \begin{bmatrix} X \\ Y \\ Z \end{bmatrix} + \begin{bmatrix} t_1 \\ t_2 \\ t_3 \end{bmatrix} \quad (8)$$

Meanwhile, with a focal length $f$, a perspective projection model can be presented as

$$r = f \tan \theta \quad (9)$$

, where $\theta$ is the angle between the incident ray and the optical axis, and $r$ is denoted as the distance from the principal point on the image frame to the projection point. Additionally, the fisheye distortion model can be approximated as:

$$r(\theta) = \sum_{i=1}^{n} k_i \theta^{2i-1}, n = 1, 2, 3, 4, \ldots \quad (10)$$

and

$$r = \sqrt{(\tfrac{X_c}{Z_c})^2 + (\tfrac{Y_c}{Z_c})^2} = \sqrt{x^2 + y^2} \quad (11)$$

To be consistent with the previous work, the $n$ was chosen to be 5 [5]. The distorted $\mathbb{P}_d := \{x_d, y_d\} \in \mathbb{R}^2$ is a nonlinear refraction by the fisheye lens onto $\mathbb{P} := \{x, y\} \in \mathbb{R}^2$, which is the normalized camera coordinate described in the equation (5).

Given a normal perspective image, we could obtain the pixel coordinates $(u, v)$ easily. In order to operate adding distortion onto this normal perspective image, we need to obtain $\theta$ in equation (3). We did a similarity transformation upon equation (5) using the pixel coordinates information.

$$r = \frac{\sqrt{(u-c_x)^2 + (v-c_y)^2}}{f} \quad (12)$$

, and by using the relationship in equation (3), we can have

$$\theta = \arctan(\frac{\sqrt{(u-c_x)^2 + (v-c_y)^2}}{f^2}) \quad (13)$$

.

From here, the distorted $r_d(\theta)$ can be obtained by adding reasonable distortion parameters $k_1, k_2, k_3, k_4, k_5$ using equation (4) and the correspondence between $\mathbb{P}_d$ and $\mathbb{P}$ can be expressed as

$$P_d = r_d(\cos\phi, \sin\phi)^T, \phi = \arctan(\frac{u-c_x}{v-c_y}) \tag{14}$$

The $\phi$ indicates the angle between the rag connects the projected points and the center of image plane and the *x*-axis of the image coordinate system. By following the previous works' assumption that the pixel coordinate system is orthogonal, we can get a new distorted pixel coordinates $(u_d, v_d)$ converted by the distorted $P_d$ as

$$\begin{bmatrix} u_d \\ v_d \end{bmatrix} = \begin{bmatrix} f_x & 0 \\ 0 & f_y \end{bmatrix} \begin{bmatrix} x_d \\ y_d \end{bmatrix} \tag{15}$$

In summary, by going through the processes described above, we are going to accurately estimate the needed 9 parameters $k_1, k_2, k_3, k_4, k_5, f_x, f_y, c_x, c_y$ to synthesize sets of distorted fisheye images pair via the perspective/ normal wireframe dataset [21]. The fisheye- perspective images pair are used to train our W-Pix2PixGAN model, whereas the parameters used are used to sequentially train the parameter- calibration module as the label.